\documentclass{article}
\usepackage{ijcai25}

\usepackage{times}
\usepackage{soul}
\usepackage{url}
\usepackage[hidelinks]{hyperref}
\usepackage[utf8]{inputenc}
\usepackage[small]{caption}
\usepackage{graphicx}
\usepackage{amsmath}
\usepackage{amsthm}
\usepackage{booktabs}
\usepackage{algorithm}
\usepackage{algorithmic}
\usepackage[switch]{lineno}
\usepackage{xcolor}
\usepackage{amssymb}
\usepackage{enumitem}

\title{MTPNet: Multi-Grained Target Perception for Unified Activity Cliff Prediction}

\author{
Zishan Shu$^{1,3}$\footnote{Equal Contribution.}
\and
Yufan Deng$^{1,3}$$^*$\and
Hongyu Zhang$^{1,3}$$^*$\and
Zhiwei Nie$^{1,2,3}$$^\dag$ \And
Jie Chen$^{1,2,3}$\footnote{Corresponding author.}
\\
\affiliations
$^1$School of Electronic and Computer Engineering, Peking University, Shenzhen, China\\
$^2$Pengcheng Laboratory, Shenzhen, China\\
$^3$AI for Science (AI4S)-Preferred Program, Peking University Shenzhen Graduate School, China\\
\emails
\{zishanshu, dengyufan10, zhanghy\}@stu.pku.edu.cn,\\
\{zhiweiNie, jiechen2019\}@pku.edu.cn,
}

\begin{document}

\maketitle

\begin{abstract}

Activity cliff prediction is a critical task in drug discovery and material design. Existing computational methods are limited to handling single binding targets, which restricts the applicability of these prediction models. In this paper, we present the Multi-Grained Target Perception network (MTPNet) to incorporate the prior knowledge of interactions between the molecules and their target proteins. Specifically, MTPNet is a unified framework for activity cliff prediction, which consists of two components: Macro-level Target Semantic (MTS) guidance and Micro-level Pocket Semantic (MPS) guidance. By this way, MTPNet dynamically optimizes molecular representations through multi-grained protein semantic conditions. To our knowledge, it is the first time to employ the receptor proteins as guiding information to effectively capture critical interaction details. Extensive experiments on 30 representative activity cliff datasets demonstrate that MTPNet significantly outperforms previous approaches, achieving an average RMSE improvement of 18.95\% on top of several mainstream GNN architectures.
Overall, MTPNet internalizes interaction patterns through conditional deep learning to achieve unified predictions of activity cliffs, helping to accelerate compound optimization and design. Codes are available at: https://github.com/ZishanShu/MTPNet.
    
\end{abstract}

\section{Introduction}
\begin{figure}[ht]
    \centering
    \includegraphics[width=\linewidth]{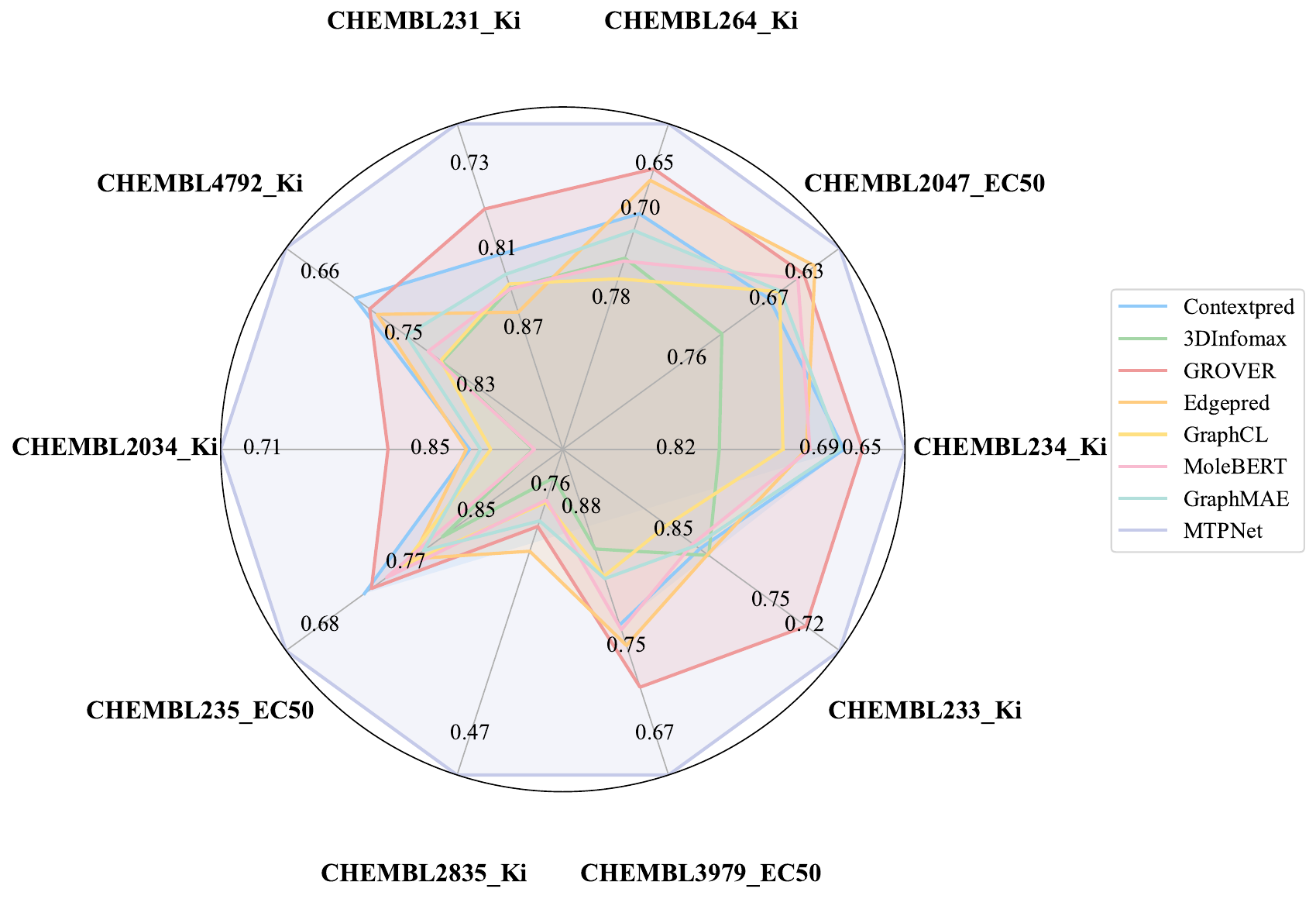} 
    \caption{The overall RMSE performance comparison of various mainstream models across multiple activity cliff datasets. The radar chart shows that MTPNet significantly outperforms existing methods, achieving a 7.2\% improvement over current SOTA models, highlighting the critical role of incorporating receptor protein information in enhancing activity cliff prediction performance. RMSE values are shown in reciprocal form to facilitate presentation.}
    \label{fig:pic/Experiment_1}
    \vspace{-2ex}
\end{figure}

In the field of drug discovery and design, Activity Cliffs (AC) refer to the phenomenon where minor structural changes in molecules lead to significant differences in biological activity \cite{van2022exposing}. Studying ACs is crucial because even compounds with similar structures can exhibit drastically different biological activities, complicating the drug design process. Traditional computational methods primarily rely on molecular fingerprint comparison and similar techniques for activity cliff prediction \cite{Consonni_Todeschini_2010}, but suffer from limited robustness \cite{Wang_2017}. In recent years, GNN-based deep learning approaches have emerged as leaders in this field \cite{shin2024dynamic,meng2024towards,yang2023gpmo,zhu2023molhfhierarchicalnormalizingflow}, overcoming the limitations of traditional techniques. For instance, MoleBERT \cite{Mole-BERT2023} integrates GNNs with pre-training to improve molecular representation accuracy, significantly improving predictive performance. Additionally, GNN-based models like ACGCN \cite{Park2022ACGCN} and MolCLR \cite{Wang2022MolecularContrastive} have effectively captured complex structure-activity relationships, advancing the accuracy and applicability of activity cliff prediction.

\begin{figure}[htbp]
    \centering
    \includegraphics[width=\linewidth]{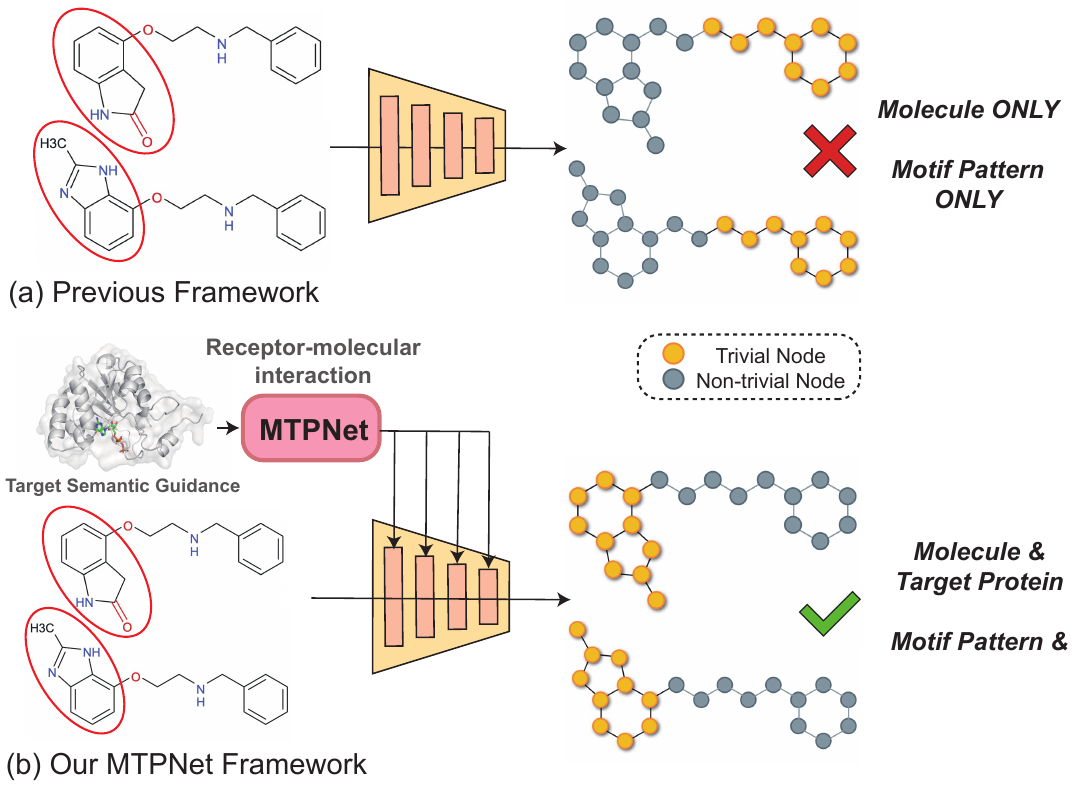} 
    \caption{ Motivation for incorporating receptor protein information in MTPNet.
    (a) The traditional methods, which only consider molecular features, focus on homogeneous and less significant parts (Motif Pattern).
    (b) MTPNet, by incorporating receptor protein information, better captures the interaction between molecules and receptors, focusing on the substitution differences in small molecules, thereby effectively revealing the underlying mechanisms of activity cliff formation.}
    \label{fig:pic/motiva}
    \vspace{-2ex}
\end{figure}

However, existing methods primarily focus on modeling molecules themselves, overlooking the critical role of paired receptor proteins in chemical reactions, particularly in the context of activity cliff (AC) prediction. As shown in Figure \ref{fig:pic/motiva}, these methods face two major challenges: First, the insufficient use of protein features hampers accurate modeling of interactions between molecules and proteins, impacting the precision of AC predictions. Second, these methods struggle to generalize across various types of AC prediction tasks, constraining their applicability to different binding targets. The latter, in particular, has become a significant bottleneck that hinders the widespread adoption of existing approaches. The fundamental principle of ACs suggests that even minor structural changes in molecules can lead to drastic shifts in biological activity, typically driven by complex interactions between ligands and receptor proteins. However, current methods fail to effectively capture these critical dynamic interaction characteristics. To address these challenges, we introduce the Multi-Grained Target Perception (MTP) module, which effectively integrates conditional information from receptor proteins and enhances the model's ability to perceive subtle structural changes. Specifically, the MTP module combines both macro and micro-level semantic guidance, enabling the identification of broad interaction patterns between molecules and the precise detection of small structural variations that result in significant differences in biological activity.

Building upon the MTP module, we further develop MTPNet, a unified framework for activity cliff prediction. MTPNet leverages the MTP module to incorporate receptor protein information into the molecule feature extraction process, enabling efficient predictions across multiple binding targets. By precisely modeling the interactions between molecules and proteins, MTPNet significantly outperforms existing models on datasets with single binding targets, demonstrating superior prediction performance and interpretability. Specifically, experimental results indicate that MTPNet improves RMSE by 7.2\% compared to existing SOTA models. Furthermore, MTPNet achieves an Area Under the Curve (AUC) of 0.924, surpassing models such as Mole-BERT (AUC = 0.902) and MolCLR (AUC = 0.896), thereby highlighting its robust generalization capabilities and practical application value across multiple receptor-ligand systems. Moreover, plug-and-play evaluations reveal that with the MTP module, PCC shows an average improvement of 11.6\%, R² improves by 17.8\%, and RMSE improves by 19.0\%, further demonstrating the module's seamless integrability and effectiveness.

In summary, the main contributions of this paper are summaried as follows:

\setlist[itemize,1]{left=0pt}

\begin{itemize}
    \item \textbf{Integration of Receptor Protein Perception:} To the best of our knowledge, MTPNet is the first conditional framework to incorporate receptor protein information into activity cliff prediction task, internalizing interaction patterns through progressive conditional deep learning.
    
    \item \textbf{Multi-Grained Target Perception Module:} We propose the Multi-Grained Target Perception (MTP) module, which dynamically optimizes molecular representations through Macro-level Target Semantic (MTS) guidance and Micro-level Pocket Semantic (MPS) guidance, effectively enhancing the predictive performance of the model by complementing the interaction patterns at different levels.
    
    \item \textbf{Unified Framework for Activity Cliff Prediction:} MTPNet provides a unified solution for activity cliff prediction across diverse receptor-ligand systems, offering applicability and interpretability.
    
    \item \textbf{Superior Predictive Performance:} Extensive experiments on 30 representative activity cliff datasets demonstrate that MTPNet achieves significant improvements in predictive performance, including an average RMSE reduction of 18.95\% on top of several mainstream GNN architectures.
\end{itemize}

\section{Related Work}
\textbf{Activity Cliff Prediction Methods:} Traditional computational methods primarily rely on techniques such as molecular fingerprint comparisons to predict Activity Cliffs \cite{Moriwaki_Tian_Kawashita_Takagi_2018}. Early computational approaches utilized traditional machine learning methods like Support Vector Machines (SVM) \cite{Vapnik2013} and Support Vector Regression (SVR) \cite{Drucker1996SVR} for Activity Cliffs prediction. Although significant progress has been made in achieving high-throughput prediction results, these methods still suffer from limitations in robustness \cite{Dong_Wang_Yao_Zhang_Cheng_Ouyang_Lu_Cao_2018}, generalization \cite{Butler_Davies_Cartwright_Isayev_Walsh_2018}, and interpretability \cite{Moriwaki_Tian_Kawashita_Takagi_2018}.

Recent advances in deep learning have significantly improved the prediction of activity cliffs, particularly through enhanced molecular feature representation \cite{NEURIPS2023_cc83e973}. Early studies, such as Iqbal et al. \cite{Iqbal2021ActivityCliffs}, utilized Convolutional Neural Networks (CNNs) to capture spatial features of molecular structures, showcasing the potential of CNNs in molecular data processing. However, as the demand for more powerful molecular embeddings grew, Graph Neural Networks (GNNs) emerged as a superior alternative due to their ability to directly model molecular graphs \cite{du2024mmgnn}, effectively capturing complex molecular interactions and learning rich embeddings \cite{xiang2024image,zheng2024cross,nie2024multi,wu2024semi,nie2024hunting,feng2025aenerfaugmentingeventbasedneural}. For instance, the MoleBERT model \cite{Mole-BERT2023}, which integrates GNNs with pretraining techniques, has significantly improved predictive performance across multiple datasets. Park et al. \cite{Park2022ACGCN} introduced the ACGCN model, addressing the information loss issues commonly associated with traditional fingerprint-based methods in structure-activity relationship (SAR) analysis. Additionally, MolCLR \cite{Wang2022MolecularContrastive} employs contrastive learning to pretrain molecular graphs, while Transformer-based models have further advanced the field. DeepAC \cite{Chen2022DeepAC}, a conditional Transformer model, leverages SMILES sequences and activity differences to predict ACs and generate novel compounds. GROVER \cite{GROVER2020}, which combines Transformer and GNN architectures, captures both local and global molecular features, achieving outstanding performance.

\textbf{Protein Language Models:} Protein language models (PLMs) have significantly advanced receptor protein feature extraction \cite{zhao2023semignnppiselfensemblingmultigraphneural,li2023glpocket,10979347}. ESM \cite{Lin2023ESM} uses self-supervised learning to pretrain on protein sequences, capturing rich features for downstream receptor function prediction, particularly on large-scale datasets \cite{wu2025rethinkingtextbasedproteinunderstanding,ZHANG2024102092}. ProteinBERT \cite{brandes2022proteinbert}, pretrained on large protein datasets, optimizes BERT for sequence modeling, showing high adaptability and efficiency. DeepProSite \cite{Fang2023DeepProSite} integrates sequence and structural information to improve receptor function prediction accuracy. DeepProtein \cite{Xie2024DeepProtein} provides a versatile library and benchmark for protein sequence learning, enhancing predictions of function, localization, and interactions. SaProt \cite{Su2023SaProt} integrates sequence and structural data for precise feature extraction, excelling in ligand binding and function prediction.

\section{Methodology}
\begin{figure*}[h]
    \centering
    \includegraphics[width=\linewidth]{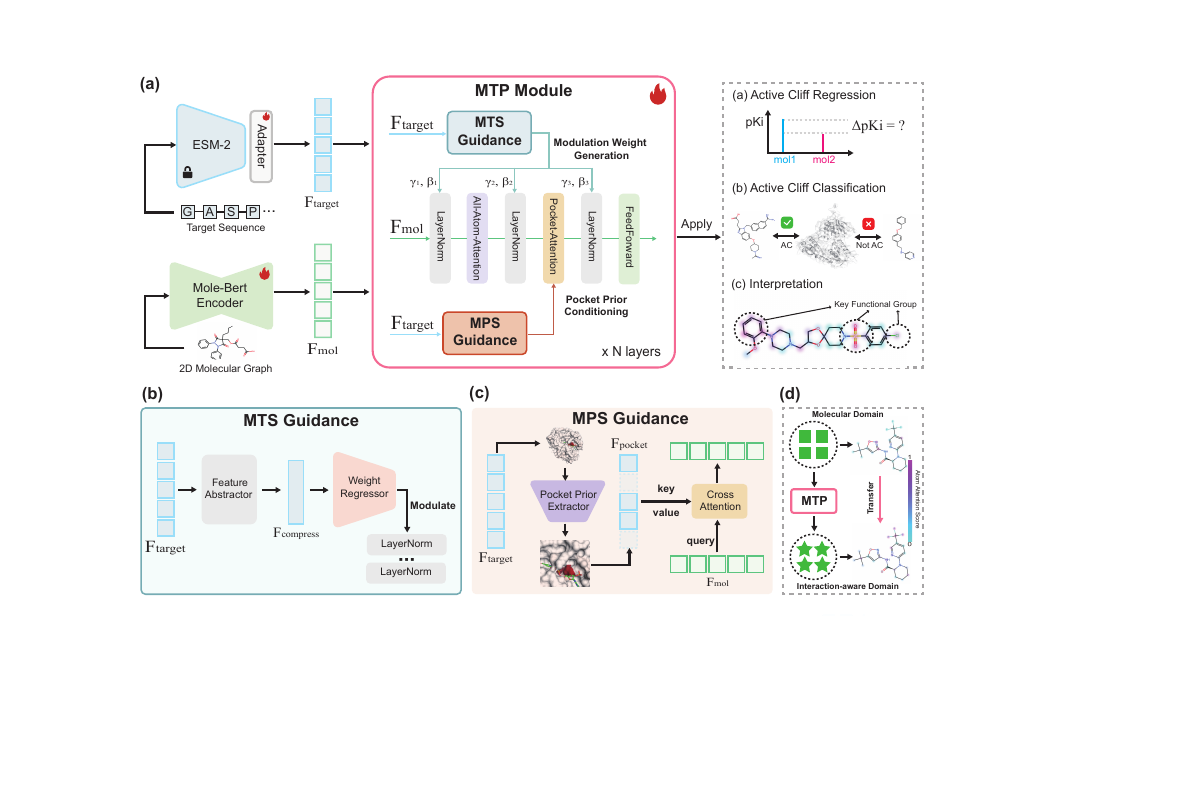} 
    \caption{The overview of MTPNet. (a) The complete workflow of MTPNet, beginning with the embedding of molecules and target proteins, followed by the MTP module, and concluding with three key applications: activity cliff regression, activity cliff classification, and molecular interpretation. (b) Detailed architecture of the Macro-level Target Semantic (MTS) Guidance, which captures global receptor semantic information. (c) Detailed architecture of the Micro-level Pocket Semantic (MPS) Guidance, which focuses on local receptor-ligand interactions within binding pockets. (d) The transformation of the molecular domain into an interaction-aware domain after passing through the MTP module, highlighting its physicochemical significance.}
    \label{fig:pic/OVERVIEW3}
    \vspace{-1ex}
\end{figure*}

\subsection{Preliminaries}

\textbf{Activity Cliff Definition:} In the process of molecular binding with receptor proteins, the binding target describes the spatial arrangement and interaction patterns between molecules and protein receptors. Different binding targets can affect how molecules bind to receptor proteins, thereby influencing the training of the model. Therefore, understanding and handling different binding targets is crucial for improving the accuracy of activity cliff prediction models. In the task of activity cliff prediction, based on the diversity of different datasets and binding targets, we categorize binding targets into single binding target and multiple binding targets.

Formally, let \( D \) be an activity cliff molecular dataset, where each \( x_i \) represents the input features of a molecular-receptor pair, and \( y_i \) represents the corresponding continuous property value, i.e., the change in compound potency values (\(\Delta pK_i\)), reflecting the difference in assay-independent equilibrium constants (\( K_i \)). The input features \( x_i \) include the receptor protein features \( x_i^{\text{pro}(m)} \) and the ligand molecule features \( x_i^{\text{mol(m)}} \), which are fused using the Multi-Grained Target Perception (MTP) Module to capture the important interaction features between the protein and ligand.

\textbf{Single Binding Target:} In single binding target learning, it is assumed that all molecular-receptor pairs follow the same binding target, and the dataset is represented as:

\begin{equation}
D = \{ (x_i^{\text{pro}}, x_i^{\text{mol}}, y_i) \}_{i=1}^N
\end{equation}

The learning objective is to fit a mapping function:

\begin{equation}
f: (x_i^{\text{pro}}, x_i^{\text{mol}}) \to y_i
\end{equation}

\textbf{Multiple Binding Targets:} In multiple binding target learning, it is assumed that the dataset contains multiple different and independent binding targets, and the dataset is represented as:

\begin{equation}
D = \bigcup_{m=1}^{M} D_m
\end{equation}

where each \( D_m = \{ (x_i^{\text{pro}(m)}, x_i^{\text{mol}(m)}, y_i^{(m)}) \}_{i=1}^{N_m} \) corresponds to the \( m \)-th binding target. The learning objective remains to fit a general mapping function:

\begin{equation}
f_m: (x_i^{\text{pro}(m)}, x_i^{\text{mol}(m)}) \to y_i^{(m)}
\end{equation}

\subsection{MTPNet}
MTPNet is a receptor-aware framework designed to unify activity cliff prediction tasks across multiple binding targets, achieving efficient and flexible modeling. At its core lies the Multi-Grained Target Perception (MTP) Module, which employs a Multigranularity Protein Semantic Condition through Macro-level Target Semantic (MTS) guidance and Micro-level Pocket Semantic (MPS) guidance to dynamically capture the complex interaction patterns between receptors and ligands. In the MTPNet framework, molecules and target proteins are first embedded, where the ESM2 model is used to extract deep semantic information from proteins, and the Mole-BERT model is employed to capture key features of molecules. Subsequently, the MTP module alternates between global context integration and local structural refinement, progressively optimizing ligand feature representations and aligning them with target features. This approach significantly enhances the capability to model activity cliff phenomena.

MTPNet leverages MTP’s global-local guidance mechanism to dynamically adapt to multiple binding targets, avoiding redundancy of separate models.
This design seamlessly integrates global features \( F_{\text{target}} \) and local features \( F_{\text{pocket}} \) into the optimization of ligand representations \( F_{\text{mol}} \), unifying the feature modeling process across multiple binding targets. By fusing receptor protein and ligand features and modeling multiple binding targets through MTP, MTPNet not only addresses the challenges of activity cliff prediction in multi-binding target scenarios but also provides a highly efficient and accurate framework for diverse receptor-ligand systems. This makes MTPNet a unified and flexible paradigm for activity cliff modeling.

\subsubsection{a. Multi-Grained Target Perception (MTP) Module}

The Multi-Grained Target Perception (MTP) Module employs a Multigranularity Protein Semantic Condition to dynamically align ligand and target features through global and localized interaction modeling. This strategy combines two complementary mechanisms: Macro-level Target Semantic (MTS) Guidance and Micro-level Pocket Semantic (MPS) Guidance. In the MTP module, the input ligand feature matrix \( F_{\text{mol}} \in \mathbb{R}^{m \times d}\) and target protein feature matrix \( F_{\text{target}} \in \mathbb{R}^{n \times d}\)are iteratively refined layer by layer through global and localized semantic modeling to optimize \( F_{\text{mol}} \).

Thus, the layer-by-layer adjustment process is essentially a recursive optimization of \( F_{\text{mol}} \), while target features \( F_{\text{target}} \) provide global and localized semantic guidance alternately through MTS and MPS. By alternating between these two mechanisms and using attention mechanisms, the MTP module dynamically captures critical molecular-receptor interaction patterns underlying activity cliffs.

In the MTP module, the outputs of MTS (see Eq.\ref{eq:MTS}) and MPS (see Eq.\ref{eq:MPS}) are alternately used within a stacked attention framework, enabling progressive updates to the ligand representation. This process is described as follows:

First, the MTS Guidance generates the initial optimized ligand features by aligning the ligand representation \( F_{\text{mol}} \) with the global context of the receptor \( F_{\text{target}} \). This alignment is achieved through a self-attention mechanism (SA) combined with conditionally normalized ligand features, as defined in Eq.\ref{eq:MTS}. The initial ligand representation is computed as:

\begin{equation} 
F_{\text{mol}}^{(0)} = \Phi_{\text{MTS}}(F_{\text{mol}}, F_{\text{target}}) 
\label{eq:init} 
\end{equation}

At the \( l \)-th layer, the ligand features are refined by alternately incorporating localized receptor information from the MPS Guidance. Using the cross-attention mechanism, the MPS Guidance integrates the ligand features \( F_{\text{mol}} \) with pocket features \( F_{\text{pocket}} \), as defined in Eq.\ref{eq:MPS}. The ligand representation at each layer is updated iteratively as:

\begin{equation} 
F_{\text{mol}}^{(l)} = F_{\text{mol}}^{(l-1)} + \Phi_{\text{MPS}}(F_{\text{mol}}^{(l-1)}, F_{\text{target}})
\end{equation}

In each iteration, the refined ligand features \( F_{\text{mol}}^{(l)} \) undergo a further transformation that includes a two-layer feedforward network with dropout and ReLU activation. This process ensures that the updated features effectively capture the dynamic receptor-ligand interaction patterns.

Finally, after \( L \) stacked layers of iterative refinement, the MTP module outputs the fully optimized ligand features \( \Phi_{\text{MTP}} \), which comprehensively integrate both global (MTS) and localized (MPS) receptor information (see Eq.\ref{eq:MTP}).

\begin{multline}
\Phi_{\text{MTP}}(F_{\text{mol}}, F_{\text{target}}) = \\
\Phi_{\text{MTS}}(F_{\text{mol}}, F_{\text{target}}) + \sum_{l=1}^{L} \Phi_{\text{MPS}}(F_{\text{mol}}^{(l-1)}, F_{\text{target}})    
\label{eq:MTP}
\end{multline}

The MTP module provides a comprehensive framework for global and localized semantic optimization. Within the multi-layer stacked attention structure, the MTP module combines the global and localized feature extraction capabilities of MTS and MPS, effectively addressing the challenges posed by multi-binding target scenarios while efficiently integrating receptor protein and ligand molecule features. This approach not only captures subtle molecular structural changes that lead to activity cliffs but also significantly enhances the performance of molecular property prediction.

\subsubsection{b. Macro-level Target Semantic (MTS) Guidance}

The Macro-level Target Semantic (MTS) Guidance aims to extract dynamic semantic information from the global context of receptor proteins to align ligand features \( F_{\text{mol}} \) with target features \( F_{\text{target}} \), facilitating the learning of the mapping \( f_m \) from \( x \) to \( y \). MTS uses the Feature Abstractor to facilitate the extraction of global semantic information from the receptor features. The Feature Abstractor compresses the receptor features to capture their key semantic characteristics. Specifically, it performs average pooling on the input target features \( F_{\text{target}} \), reducing the feature dimensions while extracting global semantic information, thus providing simplified.

\begin{equation}
F_{\text{compress}} = \text{AvgPool}(F_{\text{target}})
\end{equation}

Then, the global semantic embedding is passed through a Weight Regressor to generate dynamic conditional weights \((\gamma_i, \beta_i)\), which are used to adjust the distribution of ligand features. The Weight Regressor module is responsible for generating these weights through a linear transformation, with the core strategy being the linear transformation of global semantic embeddings. This allows the model to learn how to adjust the feature distribution at each layer. Specifically, the Weight Regressor generates a set of dynamic conditional weights based on the input global semantic information, which are then used to modulate the ligand features via Adaptive Layer Normalization (AdaLN), providing dynamic adaptation to different inputs and enhancing the model’s flexibility and performance.

\begin{equation}
\gamma_1, \beta_1, \gamma_2, \beta_2, \gamma_3, \beta_3 = \text{Linear}(F_{\text{compress}})
\end{equation}

In each layer, the ligand features \( F_{\text{mol}} \) are refined using Adaptive Layer Normalization (AdaLN), incorporating the conditional weights \(\gamma_i\) and \(\beta_i\) generated by the Weight Regressor to dynamically adjust the feature distribution:

\begin{equation}
\text{LN}_i^{\text{cond}} = \text{LN}_i (\gamma = \gamma_i, \beta = \beta_i)
\end{equation}

The conditionally normalized ligand features are passed into a self-attention (SA) module to extract global contextual representations of the ligand.

\begin{equation}
    \Phi_{\text{MTS}}(F_{\text{mol}}, F_{\text{target}}) = \text{SA}(F_{\text{mol}} \mid \text{LN}_i = \text{LN}_i^{\text{cond}})
    \label{eq:MTS}
\end{equation}

Through this process, the Macro-level Target Semantic (MTS) Guidance dynamically adjusts ligand features to align them with the global semantic context of the receptor. This alignment ensures that the ligand representation captures the global semantic patterns induced by the receptor, thereby supporting further optimization of ligand features and improving the extraction of protein-ligand interaction features.

\subsubsection{c. Micro-level Pocket Semantic (MPS) Guidance}

The Micro-level Pocket Semantic (MPS) Guidance focuses on the receptor's binding pocket region to capture local interaction patterns between receptors and ligands. By integrating the binding pocket features \( F_{\text{pocket}} \) with ligand features \( F_{\text{mol}} \), the MPS Guidance refines the mapping \( f_m \) from \( x \) to \( y \), enhancing the model's understanding of localized receptor-ligand interactions. 

Specifically, for the target features \( F_{\text{target}} \in \mathbb{R}^{n \times d} \) and ligand features \( F_{\text{mol}} \in \mathbb{R}^{m \times d} \), the binding pocket features \( F_{\text{pocket}} \in \mathbb{R}^{p \times d} \) are extracted from \( F_{\text{target}} \) using Pocket Prior Extractor (e.g., Cavity Plus \cite{xu2018cavityplus}). These pocket features are then combined with the ligand features \( F_{\text{mol}} \) through a cross-attention mechanism to achieve a deep interaction. 

In the cross-attention mechanism, the molecule features \( F_{\text{mol}} \) are used to compute the query vector \( Q_{mol} \), while the ligand features \( F_{\text{mol}} \) and pocket features \( F_{\text{pocket}} \) are concatenated to generate the key vector \( K_{pocket} \) and value vector \( V_{pocket} \). The attention weights are computed using the scaled dot-product attention mechanism:

\begin{multline}
\text{Attention}(Q_{mol}, K_{pocket}, V_{pocket}) = \\
\text{Softmax}\left(\frac{Q_{mol}K_{pocket}^\top}{\sqrt{d_k}}\right)V_{pocket}
\end{multline}

where \( Q_{mol} = W_q F_{\text{mol}}, K_{pocket} = W_k F_{\text{pocket}}, V_{pocket} = W_v F_{\text{pocket}} \) and \(W_q, W_k, W_v \in \mathbb{R}^{ d \times d} \) are learnable linear transformation matrices that extract the corresponding representations from the ligand and pocket features. 

Through this process, the MPS Guidance effectively integrates ligand features \( F_{\text{mol}} \) and pocket features \( F_{\text{pocket}} \), capturing the fine-grained interaction patterns between the receptor binding pocket and the ligand. By focusing on relevant molecular substructures, it enhances the model's sensitivity to subtle changes in molecular activity.

\begin{equation}
    \Phi_{\text{MPS}}(F_{\text{mol}}, F_{\text{target}}) = \text{CrossAttention}(F_{\text{mol}}, F_{\text{pocket}})
    \label{eq:MPS}
\end{equation}

The MPS Guidance provides localized semantic alignment, complementing the global contextual representations generated by the Macro-level Target Semantic (MTS) Guidance, ensuring a comprehensive understanding of receptor-ligand interactions.






\section{Experiments}
\subsection{Data and Experimental Setups}
All evaluations in this section are conducted on datasets from MoleculeACE (Activity Cliff Estimation) \cite{van2022exposing}, an open-access benchmarking platform available on GitHub at https://github.com/molML/MoleculeACE. This platform provides over 35,000 molecules distributed across 30 macromolecular targets, each corresponding to a separate dataset. Among these, 12 datasets contain fewer than 1,000 molecules in the training set, making MoleculeACE particularly suitable for evaluating model performance in low-data regimes.

In our experiments, we use the MTPNet framework, which integrates the Multi-Grained Target Perception (MTP) Module to dynamically model receptor-ligand interactions. MTP leverages both macro-level receptor context and micro-level binding pocket features to refine ligand representations, enabling precise activity cliff estimation. This receptor-aware design ensures that MTPNet can effectively capture the complex interplay between ligands and receptors, even in scenarios with limited data.

\begin{figure*}[h]
    \centering
    \includegraphics[width=\linewidth]{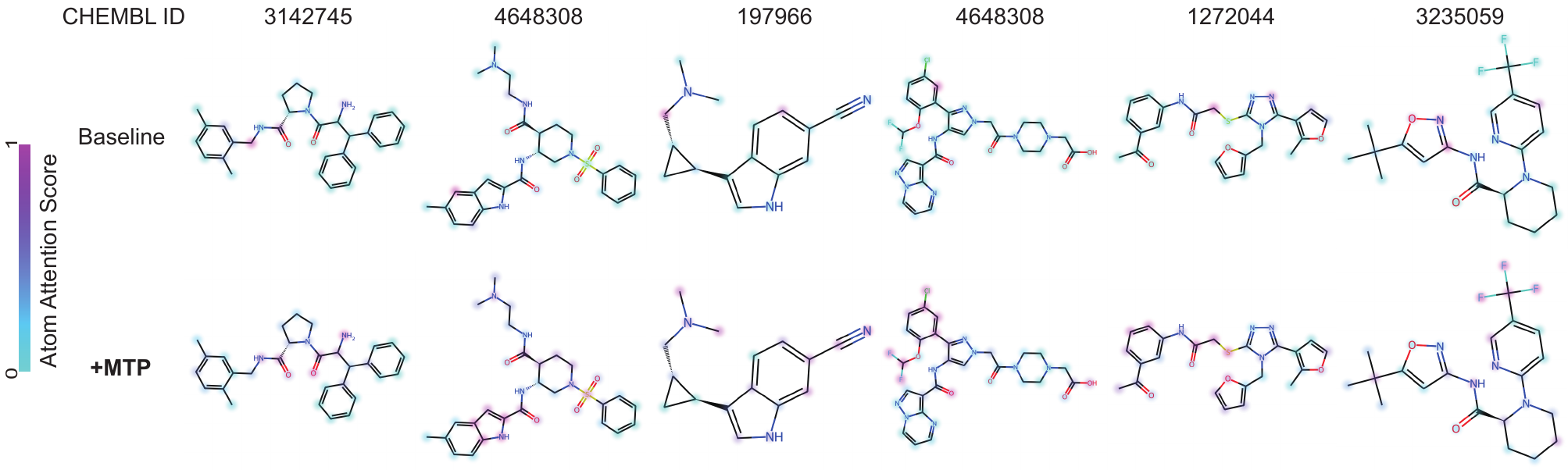} 
    \vspace{-1ex}
    \caption{Visualization of Atom Attention Scores for Molecules in CHEMBL v29. Figure \ref{fig:pic/inter} compares the atom attention distributions between the baseline model and the MTP module for six representative molecules (CHEMBL IDs: 3142745, 4648308, 197966, 4648308, 1272044, and 3235059) from the CHEMBL v29 dataset. The attention scores are color-coded, where purple indicates high attention and blue indicates low attention, as shown in the color bar on the left. The MTP module demonstrates a significantly more focused and meaningful attention distribution, effectively identifying key functional groups (e.g., amino groups, carbonyl groups) and specific chemical bonds (e.g., double bonds, triple bonds) that play critical roles in activity cliff phenomena. In contrast to the baseline model's dispersed attention patterns, the MTP module achieves more focused and precise attention, effectively capturing key molecular interactions and uncovering the chemical principles behind activity cliff phenomena.}
    \label{fig:pic/inter}
    \vspace{-1ex}
\end{figure*}

\subsection{Performance Evaluation and Comparison}

We first compared MTPNet with various machine learning (ML) and deep learning (DL) baseline algorithms on the MoleculeACE dataset, focusing on evaluating the prediction performance of activity cliff molecules. The specific results are presented in Figure \ref{fig:pic/Experiment_1} and Appendix. The analysis results indicate that MTPNet outperforms all baseline methods across the 30 activity cliff datasets, achieving the lowest RMSE values. Notably, MTPNet exhibits an average improvement of 18.95\%, which is significantly higher than the performance gains achieved by current state-of-the-art (SOTA) pretraining methods. This finding emphasizes the importance of integrating receptor protein information in effectively addressing activity cliff tasks.

\begin{table}[h]
\small
\centering
\begin{tabular}{ccccc}
\toprule
{Model} & {PCC ↑}  & {R² ↑}  & {RMSE ↓} & {Param. ↓}\\
\hline
$\text{GCN}_{\textcolor{blue}{\text{ICLR2017}}}$ & 0.711 & 0.501& 0.950& 1.11M \\
w/ Scale up & 0.737 & 0.541& 0.915& 3.17M \\
\textbf{w/ MTP} &\textbf{0.837} &\textbf{0.693} &\textbf{0.744} &\textbf{3.83M} \\
\hline
$\text{GAT}_{\textcolor{blue}{\text{ICLR2018}}}$ &0.706 &0.497 &0.956 &6.69M\\
w/ Scale up &0.715 &0.506 &0.945 &9.85M\\
\textbf{w/ MTP} &\textbf{0.828} &\textbf{0.683} &\textbf{0.756} &\textbf{9.42M} \\
\hline
$\text{GIN}_{\textcolor{blue}{\text{ICLR2019}}}$ &0.718 &0.509 &0.941 &2.25M\\
w/ Scale up &0.731 &0.535 &0.922 &5.02M\\
\textbf{w/ MTP} & \textbf{0.826} &\textbf{0.674} &\textbf{0.767} &\textbf{4.97M}  \\
\hline
$\text{GraphTrans}_{\textcolor{blue}{\text{NIPS2021}}}$& 0.719 &0.515 &0.936 &3.43M\\
w/ Scale up & 0.751 &0.559 &0.897 &6.88M\\
\textbf{w/ MTP} & \textbf{0.828} &\textbf{0.676} &\textbf{0.765} &\textbf{6.15M}\\
\hline
$\text{MolCLR}_{\textcolor{blue}{\text{NMI2023}}}$ &0.715 &0.504 &0.946 &2.04M \\
w/ Scale up &0.724 &0.526 &0.929 &4.53M \\
\textbf{w/ MTP} &\textbf{0.822} &\textbf{0.668} &\textbf{0.774} &\textbf{4.76M}\\
\hline
$\text{Mole-BERT}_{\textcolor{blue}{\text{ICLR2023}}}$ &0.721 & 0.504 &0.947 &2.34M  \\
w/ Scale up &0.742 & 0.546 &0.906 &4.78M  \\
\textbf{w/ MTP} &\textbf{0.845} &\textbf{0.703} &\textbf{0.733} &\textbf{5.07M}\\
\bottomrule
\end{tabular}
\caption{Ablation results of baseline models under layer scale-up versus MTP Module augmentation.}
\label{table:ablation_TPM1}
\vspace{-2ex}
\end{table}

To validate the robustness and effectiveness of the MTP module, we conducted an evaluation on 30 activity cliff datasets within MoleculeACE, focusing on its impact as a plug-and-play enhancement for baseline models such as GCN \cite{GCN_Kipf_Welling}, GAT \cite{velivckovic2017graph}, GIN \cite{xu2018how}, GraphTrans \cite{NEURIPS2021_6e67691b}, MolCLR \cite{Wang2022MolecularContrastive}, and Mole-BERT \cite{Mole-BERT2023}. As shown in Table \ref{table:ablation_TPM1}, integrating the MTP module resulted in consistent improvements, with PCC increasing by 11.6\%, R² by 17.8\%, and RMSE decreasing by 19.0\%. Additionally, we conducted a “scale-up” experiment by expanding the baseline models to match the parameter size of their MTP-augmented counterparts without incorporating the MTP module. The results indicate that while parameter scaling (e.g., increasing GCN's parameter size from 1.11M to 3.17M) slightly reduced RMSE from 0.950 to 0.915, the improvement was limited and still fell far short of the performance achieved by incorporating the MTP module (RMSE further reduced to 0.744). These findings demonstrate that the MTP module excels in capturing multi-level semantic information and structural nuances between molecules and receptors, far surpassing the benefits of merely increasing model size, thereby highlighting the unique strengths of the MTP module design.

In addition to these evaluations, we also assessed the performance of MTPNet in classification tasks. Specifically, we conducted experiments on the CYP3A4 dataset \cite{Rao2022} (see Table \ref{table:classific_TPM}), which includes activity cliff data of Cytochrome P450 3A4 inhibitors/substrates experimentally measured by Veith et al. (2009) \cite{Veith2009}, comprising 3,626 active inhibitors/substrates and 5,496 inactive compounds. The results show that the MTPNet achieved an AUC of 0.924, significantly outperforming baseline models like Mole-BERT (AUC = 0.902) and MolCLR (AUC = 0.896). These results demonstrate that MTPNet excels not only in regression tasks but also in classification tasks, effectively capturing activity cliffs and showcasing its potential for drug discovery and molecular activity prediction.

\begin{table}[h]
\small
\centering
\begin{tabular}{cc}
\toprule
{Model} & {AUC}  \\
\midrule
$\text{GCN}_{\textcolor{blue}{\text{ICLR2017}}}$ &0.766\\
$\text{GAT}_{\textcolor{blue}{\text{ICLR2018}}}$ &0.773\\
$\text{SemiMol}_{\textcolor{blue}{\text{IJCAI2024}}}$  &0.857\\
$\text{GraphTrans}_{\textcolor{blue}{\text{NIPS2021}}}$ &0.890\\
$\text{Mole-BERT}_{\textcolor{blue}{\text{ICLR2023}}}$ &0.902 \\
$\text{MolCLR}_{\textcolor{blue}{\text{NMI2023}}}$ &0.896 \\
\textbf{Ours} & \textbf{0.924}   \\
\bottomrule
\end{tabular}
\caption{Comparison experiments on CYP3A4 dataset.}
\label{table:classific_TPM}
\end{table}

\subsection{Ablation Study}
To evaluate the contribution of each component in MTPNet, we conducted an ablation study (see Table \ref{table:ablation_TPM}) comparing model performance with and without the Adaptive LayerNorm (AdaLN) and Cross Attention (CA) modules. Removing AdaLN is equivalent to disabling Macro-level Target Semantic (MTS) guidance, while removing CA corresponds to disabling Micro-level Pocket Semantic (MPS) guidance. The results show that removing either the AdaLN or CA module leads to a decline in PCC, R², and RMSE, which underscores the critical role of MTS guidance and MPS guidance in capturing global receptor semantics and localized receptor–ligand interactions.

\begin{table}[h]
\small
\centering
\begin{tabular}{cccc}
\toprule
{Model} & {PCC ↑}  & {R² ↑}  & {RMSE ↓} \\
\midrule
w/o MTS \& MPS &0.737 &0.526 &0.917  \\
w/o MTS &0.830 &0.680 &0.760\\
w/o MPS &0.826 &0.676 &0.766\\
\textbf{Ours} &\textbf{0.845} &\textbf{0.703} &\textbf{0.733}\\
\bottomrule
\end{tabular}
\caption{Ablation results on the MTP Module sub-components (MTS and MPS).}
\label{table:ablation_TPM}
\vspace{-2ex}
\end{table}

\subsection{Interpretation}
MTPNet, by integrating global and local semantic guidance, accurately captures the interaction patterns between key functional groups and chemical bonds within molecules, thereby uncovering the chemical essence of activity cliff phenomena. Figure \ref{fig:pic/inter} shows that the MTP module assigns significant attention to key functional groups such as amino groups (NH2), carbonyl groups (C=O), sulfonyl groups (O=S=O), carboxyl groups (COOH), and halogen groups, as well as to specific chemical bonds such as double and triple bonds. Its attention distribution is far superior to the dispersed attention of baseline models. These functional groups and chemical bonds play critical roles in molecular binding affinity and activity changes with receptors. For instance, amino groups influence molecular activity by forming hydrogen bonds, sulfonyl groups regulate molecular solubility and receptor-binding stability due to their strong polarity, and halogen groups and carboxyl groups significantly impact molecular behavior by modulating hydrophobicity and acidity, respectively. For molecules containing both key functional groups and specific chemical bonds, the MTP module generally assigns greater attention to functional groups than to chemical bonds. This observation supports an important principle in chemistry: functional groups are more critical than chemical bonds in determining molecular properties and reactivity. By capturing this pattern, the MTP module demonstrates strong interpretability, even reflecting fundamental chemical principles. In summary, the MTP module significantly enhances the interpretability of activity cliff predictions, accurately identifying key activity sites within molecules and revealing the core influence of functional groups and chemical bonds on molecular behavior. This has profound implications for the study of chemical reaction mechanisms and protein-ligand binding rules, while also offering novel perspectives and approaches for understanding the causes of activity cliffs and exploring complex receptor-ligand interaction mechanisms.

\section{Conclusion}

By introducing the interactions of molecules and their target proteins as the guidance, MTPNet achieves unified prediction across diverse downstream tasks related to activity cliffs.
Through internalizing interaction patterns at different granularities, MTPNet comprehensively outperforms other methods on 30 representative datasets.
In addition, when MTPNet is adopted as a plug-in, the prediction performance of multiple mainstream GNN architectures is significantly improved, showing ideal usability and robustness.
Looking to the future, MTPNet is expected to achieve more efficient hit-to-lead optimization to accelerate drug design.

\appendix

\section*{Acknowledgments}
This work was supported in part by the Shenzhen Medical Research Funds in China (No. B2302037), Natural Science Foundation of China (No. 61972217, 32071459, 62176249, 62006133, 62271465),   and AI for Science (AI4S)-Preferred Program, Peking University Shenzhen Graduate School, China.

\bibliographystyle{named}
\bibliography{ijcai25}

\begin{thebibliography}{}

\bibitem[\protect\citeauthoryear{Brandes \bgroup \em et al.\egroup }{2022}]{brandes2022proteinbert}
Nadav Brandes, Dan Ofer, Yam Peleg, Nadav Rappoport, and Michal Linial.
\newblock Proteinbert: a universal deep-learning model of protein sequence and function.
\newblock {\em Bioinformatics}, 38(8):2102--2110, 2022.

\bibitem[\protect\citeauthoryear{Butler \bgroup \em et al.\egroup }{2018}]{Butler_Davies_Cartwright_Isayev_Walsh_2018}
Keith~T. Butler, Daniel~W. Davies, Hugh Cartwright, Olexandr Isayev, and Aron Walsh.
\newblock Machine learning for molecular and materials science.
\newblock {\em Nature}, page 547–555, Jul 2018.

\bibitem[\protect\citeauthoryear{Chen \bgroup \em et al.\egroup }{2022}]{Chen2022DeepAC}
Hengwei Chen, Martin Vogt, and Jürgen Bajorath.
\newblock Deepac – conditional transformer-based chemical language model for the prediction of activity cliffs formed by bioactive compounds.
\newblock {\em Digital Discovery}, 1(6):898--909, 2022.

\bibitem[\protect\citeauthoryear{Consonni and Todeschini}{2010}]{Consonni_Todeschini_2010}
Viviana Consonni and Roberto Todeschini.
\newblock {\em Molecular Descriptors}, pages 29--102.
\newblock Jan 2010.

\bibitem[\protect\citeauthoryear{Dong \bgroup \em et al.\egroup }{2018}]{Dong_Wang_Yao_Zhang_Cheng_Ouyang_Lu_Cao_2018}
Jie Dong, Ning-Ning Wang, Zhi-Jiang Yao, Lin Zhang, Yan Cheng, Defang Ouyang, Ai-Ping Lu, and Dong-Sheng Cao.
\newblock Admetlab: a platform for systematic admet evaluation based on a comprehensively collected admet database.
\newblock {\em Journal of Cheminformatics}, 10(1), Dec 2018.

\bibitem[\protect\citeauthoryear{Drucker \bgroup \em et al.\egroup }{1996}]{Drucker1996SVR}
H.~Drucker, C.~J. Burges, L.~Kaufman, A.~Smola, and V.~Vapnik.
\newblock Support vector regression machines.
\newblock In {\em Advances in Neural Information Processing Systems}, volume~9, 1996.

\bibitem[\protect\citeauthoryear{Du \bgroup \em et al.\egroup }{2024}]{du2024mmgnn}
Wenjie Du, Shuai Zhang, Jun~Xia Di~Wu, Ziyuan Zhao, Junfeng Fang, and Yang Wang.
\newblock Mmgnn: A molecular merged graph neural network for explainable solvation free energy prediction.
\newblock In {\em Proceedings of the Thirty-Third International Joint Conference on Artificial Intelligence}, pages 5808--5816, 2024.

\bibitem[\protect\citeauthoryear{Fang \bgroup \em et al.\egroup }{2023}]{Fang2023DeepProSite}
Y.~Fang, Y.~Jiang, L.~Wei, Q.~Ma, Z.~Ren, Q.~Yuan, and D.~Q. Wei.
\newblock Deepprosite: structure-aware protein binding site prediction using esmfold and pretrained language model.
\newblock {\em Bioinformatics}, 39(12):btad718, Dec 2023.

\bibitem[\protect\citeauthoryear{Feng \bgroup \em et al.\egroup }{2025}]{feng2025aenerfaugmentingeventbasedneural}
Chaoran Feng, Wangbo Yu, Xinhua Cheng, Zhenyu Tang, Junwu Zhang, Li~Yuan, and Yonghong Tian.
\newblock Ae-nerf: Augmenting event-based neural radiance fields for non-ideal conditions and larger scene, 2025.

\bibitem[\protect\citeauthoryear{Iqbal \bgroup \em et al.\egroup }{2021}]{Iqbal2021ActivityCliffs}
J.~Iqbal, M.~Vogt, and J.~Bajorath.
\newblock Prediction of activity cliffs on the basis of images using convolutional neural networks.
\newblock {\em Journal of Computer-Aided Molecular Design}, pages 1--8, 2021.

\bibitem[\protect\citeauthoryear{Kipf and Welling}{2017}]{GCN_Kipf_Welling}
Thomas~N. Kipf and Max Welling.
\newblock Semi-supervised classification with graph convolutional networks.
\newblock In {\em International Conference on Learning Representations}, 2017.

\bibitem[\protect\citeauthoryear{Li \bgroup \em et al.\egroup }{2023}]{li2023glpocket}
Peiying Li, Yongchang Liu, Shikui Tu, and Lei Xu.
\newblock Glpocket: A multi-scale representation learning approach for protein binding site prediction.
\newblock In {\em IJCAI}, pages 4821--4828, 2023.

\bibitem[\protect\citeauthoryear{Lin \bgroup \em et al.\egroup }{2023}]{Lin2023ESM}
Zeming Lin, Halil Akin, Roshan Rao, Brian Hie, Zhongkai Zhu, Wenting Lu, Nikita Smetanin, Robert Verkuil, Ori Kabeli, Yaniv Shmueli, Allan dos Santos~Costa, Maryam Fazel-Zarandi, Tom Sercu, Salvatore Candido, and Alexander Rives.
\newblock Evolutionary-scale prediction of atomic-level protein structure with a language model.
\newblock {\em Science}, 379(6637):1123--1130, 2023.

\bibitem[\protect\citeauthoryear{Lv \bgroup \em et al.\egroup }{2025}]{10979347}
Liuzhenghao Lv, Zongying Lin, Hao Li, Yuyang Liu, Jiaxi Cui, Calvin Yu-Chian Chen, Li~Yuan, and Yonghong Tian.
\newblock Prollama: A protein large language model for multi-task protein language processing.
\newblock {\em IEEE Transactions on Artificial Intelligence}, 2025.

\bibitem[\protect\citeauthoryear{Meng \bgroup \em et al.\egroup }{2024}]{meng2024towards}
Ziqiao Meng, Liang Zeng, Zixing Song, Tingyang Xu, Peilin Zhao, and Irwin King.
\newblock Towards geometric normalization techniques in se (3) equivariant graph neural networks for physical dynamics simulations.
\newblock In {\em Proceedings of the Thirty-Third International Joint Conference on Artificial Intelligence}, pages 5981--5989, 2024.

\bibitem[\protect\citeauthoryear{Moriwaki \bgroup \em et al.\egroup }{2018}]{Moriwaki_Tian_Kawashita_Takagi_2018}
Hirotomo Moriwaki, Yu-Shi Tian, Norihito Kawashita, and Tatsuya Takagi.
\newblock Mordred: a molecular descriptor calculator.
\newblock {\em Journal of Cheminformatics}, Dec 2018.

\bibitem[\protect\citeauthoryear{Nie \bgroup \em et al.\egroup }{2024a}]{nie2024hunting}
Zhiwei Nie, Daixi Li, Jie Chen, Fan Xu, Yutian Liu, Jie Fu, Xudong Liu, Zhennan Wang, Yiming Ma, Kai Wang, et~al.
\newblock Hunting for peptide binders of specific targets with data-centric generative language models.
\newblock {\em bioRxiv}, pages 2023--12, 2024.

\bibitem[\protect\citeauthoryear{Nie \bgroup \em et al.\egroup }{2024b}]{nie2024multi}
Zhiwei Nie, Hongyu Zhang, Hao Jiang, Yutian Liu, Xiansong Huang, Fan Xu, Yonghong Tian, Jie Chen, and Wen-Bin Zhang.
\newblock Multi-purpose enzyme-substrate interaction prediction with progressive conditional deep learning.
\newblock 2024.

\bibitem[\protect\citeauthoryear{Park \bgroup \em et al.\egroup }{2022}]{Park2022ACGCN}
Junhui Park, Gaeun Sung, SeungHyun Lee, SeungHo Kang, and ChunKyun Park.
\newblock Acgcn: Graph convolutional networks for activity cliff prediction between matched molecular pairs.
\newblock {\em Journal of Chemical Information and Modeling}, 62(10):2341--2351, 2022.

\bibitem[\protect\citeauthoryear{Rao \bgroup \em et al.\egroup }{2022}]{Rao2022}
Jiahua Rao, Shuangjia Zheng, Yutong Lu, and Yuedong Yang.
\newblock Quantitative evaluation of explainable graph neural networks for molecular property prediction.
\newblock {\em Patterns}, 3(12):100628, 2022.

\bibitem[\protect\citeauthoryear{Rong \bgroup \em et al.\egroup }{2020}]{GROVER2020}
Yu~Rong, Yatao Bian, Tingyang Xu, Weiyang Xie, Ying Wei, Wenbing Huang, and Junzhou Huang.
\newblock Self-supervised graph transformer on large-scale molecular data.
\newblock {\em arXiv: Biomolecules,arXiv: Biomolecules}, Jun 2020.

\bibitem[\protect\citeauthoryear{Shin \bgroup \em et al.\egroup }{2024}]{shin2024dynamic}
Dong-Hee Shin, Young-Han Son, Deok-Joong Lee, Ji-Wung Han, and Tae-Eui Kam.
\newblock Dynamic many-objective molecular optimization: Unfolding complexity with objective decomposition and progressive optimization.
\newblock In {\em Proceedings of the Thirty-Third International Joint Conference on Artificial Intelligence (IJCAI)}, pages 6026--6034, 2024.

\bibitem[\protect\citeauthoryear{Su \bgroup \em et al.\egroup }{2023}]{Su2023SaProt}
Jin Su, Chenchen Han, Yuyang Zhou, Junjie Shan, Xibin Zhou, and Fajie Yuan.
\newblock Saprot: Protein language modeling with structure-aware vocabulary, Oct 2023.
\newblock bioRxiv 2023.10.01.560349.

\bibitem[\protect\citeauthoryear{Van~Tilborg \bgroup \em et al.\egroup }{2022}]{van2022exposing}
Derek Van~Tilborg, Alisa Alenicheva, and Francesca Grisoni.
\newblock Exposing the limitations of molecular machine learning with activity cliffs.
\newblock {\em Journal of chemical information and modeling}, 62(23):5938--5951, 2022.

\bibitem[\protect\citeauthoryear{Vapnik}{2013}]{Vapnik2013}
Vladimir Vapnik.
\newblock {\em The Nature of Statistical Learning Theory}.
\newblock Springer Science \& Business Media, 2013.

\bibitem[\protect\citeauthoryear{Veith \bgroup \em et al.\egroup }{2009}]{Veith2009}
Henrike Veith, Noel Southall, Ruili Huang, Tim James, Darren Fayne, Natalia Artemenko, Min Shen, James Inglese, Christopher~P. Austin, David~G. Lloyd, and et~al.
\newblock Comprehensive characterization of cytochrome p450 isozyme selectivity across chemical libraries.
\newblock {\em Nature Biotechnology}, 27(11):1050--1055, 2009.

\bibitem[\protect\citeauthoryear{Veli{\v{c}}kovi{\'c} \bgroup \em et al.\egroup }{2017}]{velivckovic2017graph}
Petar Veli{\v{c}}kovi{\'c}, Guillem Cucurull, Arantxa Casanova, Adriana Romero, Pietro Lio, and Yoshua Bengio.
\newblock Graph attention networks.
\newblock {\em arXiv preprint arXiv:1710.10903}, 2017.

\bibitem[\protect\citeauthoryear{Wang \bgroup \em et al.\egroup }{2017}]{Wang_2017}
Yanli Wang, Stephen~H. Bryant, Tiejun Cheng, Jiyao Wang, Asta Gindulyte, Benjamin~A. Shoemaker, Paul~A. Thiessen, Siqian He, and Jian Zhang.
\newblock Pubchem bioassay: 2017 update.
\newblock {\em Nucleic Acids Research}, page D955–D963, Jan 2017.

\bibitem[\protect\citeauthoryear{Wang \bgroup \em et al.\egroup }{2022}]{Wang2022MolecularContrastive}
Y.~Wang, J.~Wang, Z.~Cao, et~al.
\newblock Molecular contrastive learning of representations via graph neural networks.
\newblock {\em Nature Machine Intelligence}, 4:279--287, 2022.

\bibitem[\protect\citeauthoryear{Wu \bgroup \em et al.\egroup }{2021}]{NEURIPS2021_6e67691b}
Zhanghao Wu, Paras Jain, Matthew Wright, Azalia Mirhoseini, Joseph~E Gonzalez, and Ion Stoica.
\newblock Representing long-range context for graph neural networks with global attention.
\newblock In M.~Ranzato, A.~Beygelzimer, Y.~Dauphin, P.S. Liang, and J.~Wortman Vaughan, editors, {\em Advances in Neural Information Processing Systems}, volume~34, pages 13266--13279. Curran Associates, Inc., 2021.

\bibitem[\protect\citeauthoryear{Wu \bgroup \em et al.\egroup }{2025}]{wu2025rethinkingtextbasedproteinunderstanding}
Juntong Wu, Zijing Liu, He~Cao, Hao Li, Bin Feng, Zishan Shu, Ke~Yu, Li~Yuan, and Yu~Li.
\newblock Rethinking text-based protein understanding: Retrieval or llm?, 2025.

\bibitem[\protect\citeauthoryear{Wu}{2024}]{wu2024semi}
Fang Wu.
\newblock A semi-supervised molecular learning framework for activity cliff estimation.
\newblock In {\em Proceedings of the Thirty-Third International Joint Conference on Artificial Intelligence}, pages 6080--6088, 2024.

\bibitem[\protect\citeauthoryear{Xia \bgroup \em et al.\egroup }{2023a}]{NEURIPS2023_cc83e973}
Jun Xia, Lecheng Zhang, Xiao Zhu, Yue Liu, Zhangyang Gao, Bozhen Hu, Cheng Tan, Jiangbin Zheng, Siyuan Li, and Stan~Z. Li.
\newblock Understanding the limitations of deep models for molecular property prediction: Insights and solutions.
\newblock In A.~Oh, T.~Naumann, A.~Globerson, K.~Saenko, M.~Hardt, and S.~Levine, editors, {\em Advances in Neural Information Processing Systems}, volume~36, pages 64774--64792. Curran Associates, Inc., 2023.

\bibitem[\protect\citeauthoryear{Xia \bgroup \em et al.\egroup }{2023b}]{Mole-BERT2023}
Jun Xia, Chengshuai Zhao, Bozhen Hu, Zhangyang Gao, Cheng Tan, Yue Liu, Siyuan Li, and Stan~Z. Li.
\newblock Mole-bert: Rethinking pre-training graph neural networks for molecules.
\newblock In {\em The Eleventh International Conference on Learning Representations, ICLR 2023, Kigali, Rwanda, May 1-5, 2023}. OpenReview.net, 2023.

\bibitem[\protect\citeauthoryear{Xiang \bgroup \em et al.\egroup }{2024}]{xiang2024image}
Hongxin Xiang, Shuting Jin, Jun Xia, Man Zhou, Jianmin Wang, Li~Zeng, and Xiangxiang Zeng.
\newblock An image-enhanced molecular graph representation learning framework.
\newblock In {\em Proceedings of the Thirty-Third International Joint Conference on Artificial Intelligence}, pages 6107--6115, 2024.

\bibitem[\protect\citeauthoryear{Xie \bgroup \em et al.\egroup }{2024}]{Xie2024DeepProtein}
Jiaqing Xie, Yue Zhao, and Tianfan Fu.
\newblock Deepprotein: Deep learning library and benchmark for protein sequence learning, 2024.

\bibitem[\protect\citeauthoryear{Xu \bgroup \em et al.\egroup }{2018}]{xu2018cavityplus}
Y~Xu, S~Wang, Q~Hu, S~Gao, X~Ma, W~Zhang, Y~Shen, F~Chen, L~Lai, and J~Pei.
\newblock Cavityplus: a web server for protein cavity detection with pharmacophore modelling, allosteric site identification and covalent ligand binding ability prediction.
\newblock {\em Nucleic Acids Research}, 46(W1):W374--W379, 2018.

\bibitem[\protect\citeauthoryear{Xu \bgroup \em et al.\egroup }{2019}]{xu2018how}
Keyulu Xu, Weihua Hu, Jure Leskovec, and Stefanie Jegelka.
\newblock How powerful are graph neural networks?
\newblock In {\em International Conference on Learning Representations}, 2019.

\bibitem[\protect\citeauthoryear{Yang \bgroup \em et al.\egroup }{2023}]{yang2023gpmo}
Xixi Yang, Li~Fu, Yafeng Deng, Yuansheng Liu, Dongsheng Cao, and Xiangxiang Zeng.
\newblock Gpmo: Gradient perturbation-based contrastive learning for molecule optimization.
\newblock In {\em IJCAI}, pages 4940--4948, 2023.

\bibitem[\protect\citeauthoryear{Zhang \bgroup \em et al.\egroup }{2024}]{ZHANG2024102092}
Haohui Zhang, Juntong Wu, Shichao Liu, and Shen Han.
\newblock A pre-trained multi-representation fusion network for molecular property prediction.
\newblock {\em Information Fusion}, 103:102092, 2024.

\bibitem[\protect\citeauthoryear{Zhao \bgroup \em et al.\egroup }{2023}]{zhao2023semignnppiselfensemblingmultigraphneural}
Ziyuan Zhao, Peisheng Qian, Xulei Yang, Zeng Zeng, Cuntai Guan, Wai~Leong Tam, and Xiaoli Li.
\newblock Semignn-ppi: Self-ensembling multi-graph neural network for efficient and generalizable protein-protein interaction prediction, 2023.

\bibitem[\protect\citeauthoryear{Zheng \bgroup \em et al.\egroup }{2024}]{zheng2024cross}
Yan Zheng, Song Wu, Junyu Lin, Yazhou Ren, Jing He, Xiaorong Pu, and Lifang He.
\newblock Cross-view contrastive fusion for enhanced molecular property prediction.
\newblock In {\em Proccedings of the Thirty-Third International Joint Conference on Artificial Intelligence}, 2024.

\bibitem[\protect\citeauthoryear{Zhu \bgroup \em et al.\egroup }{2023}]{zhu2023molhfhierarchicalnormalizingflow}
Yiheng Zhu, Zhenqiu Ouyang, Ben Liao, Jialu Wu, Yixuan Wu, Chang-Yu Hsieh, Tingjun Hou, and Jian Wu.
\newblock Molhf: A hierarchical normalizing flow for molecular graph generation, 2023.

\end{thebibliography}

\end{document}